\documentclass[accepted]{uai2025}
\usepackage{amsmath,amssymb,amsthm}
\usepackage{mathtools}
\usepackage{bm}
\usepackage{natbib}

\def\ind{\perp\!\!\!\perp}

\newcommand{\cov}{\text{cov}}

\newcommand{\Pb}{\mathbb{P}}

\newtheorem{theorem}{Theorem}

\DeclareUnicodeCharacter{03C0}{\ensuremath{\pi}}

\newcommand{\E}{\mathbb{E}}

\DeclareSymbolFont{bbold}{U}{bbold}{m}{n}
\DeclareSymbolFontAlphabet{\mathbbold}{bbold}

\usepackage{cancel}
\newtheorem{proposition}{Proposition}
\theoremstyle{definition}

\theoremstyle{remark}

\usepackage{algorithmic}
\usepackage[linesnumbered,ruled]{algorithm2e}
                        
\usepackage[american]{babel}

\usepackage{natbib} 
\usepackage{mathtools} 
\usepackage{booktabs} 
\usepackage{tikz} 

\title{Dependent Randomized Rounding for Budget Constrained Experimental Design}

\author[1]{\href{mailto:<jj@example.edu>?Subject=Your UAI 2025 paper}{Khurram~Yamin}{}}
\author[1]{Edward~Kennedy}
\author[1]{Bryan~Wilder}
\affil[1]{%
    Carnegie Mellon University\\
    Pittsburgh, Pennsylvania, USA
}

\begin{document}
\maketitle

\begin{abstract}
  Policymakers in resource-constrained settings require experimental designs that satisfy strict budget limits while ensuring precise estimation of treatment effects. We propose a framework that applies a dependent randomized rounding procedure to convert  assignment probabilities into binary treatment decisions. Our proposed solution preserves the marginal treatment probabilities while inducing negative correlations among assignments, leading to improved estimator precision through variance reduction. We establish theoretical guarantees for the inverse propensity weighted and general linear estimators, and demonstrate through empirical studies that our approach yields efficient and accurate inference under fixed budget constraints.
\end{abstract}

\section{Introduction}

Experimental design in many fields requires balancing two competing sets of objectives. On one side, practical constraints such as limited budgets or fixed resource availability demand that the total number of treated units be strictly bounded. On the other side, the design must preserve the individual-level assignment probabilities that have been determined—often based on factors such as risk or potential benefit—to ensure that the estimation of treatment effects is both statistically efficient and unbiased. In many settings, assignment probabilities for an experiment may be chosen to reflect concerns like strategy-proofness or allocating treatment to higher-need individuals. \citep{kido2023incorporatingpreferencestreatmentassignment,10.1257/aer.102.4.1279,7doi:10.1073/pnas.2008740118,wilder2024learningtreatmenteffectstreating} 

Consider, for example, a public health intervention during a flu outbreak where only a limited number of vaccines are available. In such a setting, independent Bernoulli randomization might preserve each individual’s target probability of being treated, but it does not guarantee that the total number of vaccinations will exactly match the available supply. Conversely, methods that force a deterministic count of treatments (e.g., conditional Poisson sampling) may enforce the resource constraint, yet they risk distorting the pre-specified assignment probabilities. \citep{hajek1964, Grafstrm2005ComparisonsOM} When the intended probabilities are altered, individuals who are at higher risk and thus designed to have a greater chance of receiving the vaccine may end up being underrepresented in the treated group. Such misallocation can lead to poorer health outcomes for those who need the intervention the most. Conversely, individuals with lower risk might be overtreated, potentially wasting limited resources and causing unnecessary exposure to the treatment's side effects. In both cases, the fairness and overall effectiveness of the intervention are compromised. Moreover, from a statistical standpoint, the mis-specified weights that would be used in an inverse probability weighted (IPW) estimator no longer reflect the true design, thereby inflating the variance of effect estimates \citep{hajek1964, Grafstrm2005ComparisonsOM}.

In this paper, we propose a framework based on \emph{dependent randomized rounding}—specifically, a swap rounding algorithm as introduced in \citep{chekuri2010}—that addresses these challenges. Swap rounding operates by iteratively pairing fractional assignment probabilities and redistributing probability mass between the paired units in a randomized manner. This procedure continues until an integral solution is reached. Our method exactly enforces a fixed treatment count while preserving the marginal assignment probabilities determined by the experimental design. Moreover, by inducing negative correlations among the treatment indicators, our approach leads to a reduction in the variance of standard estimators, such as the inverse probability weighted (IPW) estimator. As such, we balance the competing objectives of satisfying resource constraints while maintaining an unbiased and low-variance estimator. Furthermore, our variance reduction benefits are empirically substantial enough for small samples that our technique may be of interest just as a variance reduction technique even when budget constraints are not themselves at play. The benefits achieved in this regime from swap-rounding are somewhat surprising as despite the fact that the swap-rounding algorithm is solving an ostensibly similar problem, it was proposed for different purposes, primarily in the design of algorithms for submodular optimization \citep{chekuri2010}.  To our knowledge, no previous work has made this connection between budget-constrained experimental design and combinatorial optimization. Furthermore, while the swap-rounding algorithm had been analyzed for the purposes of optimization, its statistical properties which we explore in this paper had not previously been analyzed.

The remainder of the paper is organized as follows. We describe our dependent randomized rounding mechanism as well as the causal assumptions we make. We then discuss our strategy for integrating swap rounding into IPW estimation and derive an estimator for the variance of IPW using swap-rounding. We discuss the estimator's theoretical guarantees (unbiased estimation and lower variance when IPW is used with swap rounding than without). We then detail the unbiasedness and variance reduction properties for general linear estimators under our approach. Following this, we propose a modification to swap-rounding where we prioritize swaps between units with similar covariates. To demonstrate the efficacy of our approach in empirical studies, we use a combination of synthetic and semi-synthetic examples based on real-world data in two domains: infant health and public housing. Finally, we conclude with a discussion of the broader implications for experimental design in settings with competing constraints.

\section{Related Works}

Experimental design under resource constraints has long been a topic of interest in both survey sampling and causal inference. Classical approaches, such as conditional Poisson sampling \citep{hajek1964}, were introduced to ensure that a fixed sample size is achieved. However, these methods can distort the intended assignment probabilities and introduce bias into estimators \citep{AIRES2000133, Grafstrm2005ComparisonsOM}. Similar challenges have been observed in unequal probability sampling \citep{Isaki1984}, where the balancing of inclusion probabilities is critical to obtain reliable estimates. These limitations motivate the search for alternative methods that maintain the original probability structure while strictly enforcing allocation constraints, and that also preserve desirable properties such as unbiasedness and robustness in finite samples.

More recently, advances in combinatorial optimization have led to the development of dependent randomized rounding techniques. In particular, swap rounding \citep{chekuri2010} is a powerful method for converting fractional assignment solutions into integral ones while preserving marginal probabilities. This approach not only maintains unbiasedness by ensuring that the marginal assignment probabilities are exactly retained but also imposes a negative correlation structure among the rounded variables. Related work in submodular maximization \citep{Ageev2004} and robust approximation algorithms \citep{Srinivasan2001} further demonstrates that these rounding procedures yield solutions with enhanced concentration properties, improved stability, and robustness to model misspecification, all of which are critical for reliable inference in experimental settings. Our work builds on these ideas by employing swap rounding to enforce an exact treatment count while retaining the desired statistical properties of the design, which in turn improves the precision and decreases the variance of standard estimators like the inverse propensity weighted (IPW) and doubly robust (DR) estimators.

In addition, re-randomization methods have been explored to improve covariate balance and enhance the efficiency of experimental designs \citep{morgan2012rerandomization, Li2018rerandomization}. Re-randomization repeatedly randomizes the treatment assignments until a desired balance is achieved. Our framework, in contrast, directly enforces the resource constraint and connects with optimal allocation strategies such as Neyman allocation \citep{Neyman1934} that similarly adjusts sampling probabilities to minimize variance. This unified approach preserves the key statistical properties of the initial design and is particularly relevant in settings like public health interventions and social programs \citep{Athey2017, Imbens2015causal}, where efficient use of limited resources can have a significant social impact.

\section{Problem Formulation}

In many experimental settings, researchers are given a desired set of assignment probabilities that are designed to balance goals like welfare, efficiency, or fairness. Suppose we have a set of \( n \) candidate units. For each unit \( i \) we have a fractional assignment probability \( p_i \in [0,1] \) such that
\[
\sum_{i=1}^n p_i = B,
\]
where \( B \) represents a fixed resource constraint (for example, the exact number of vaccines available in a public health intervention). Our goal is to convert this fractional probability vector into a distribution over binary treatment assignments \( A \in \{0,1\}^n \) that meets the following criteria:

\begin{enumerate}
    \item \textbf{Feasibility:} Every assignment \( A \) satisfies the resource constraint exactly, i.e.,
    \[
    \sum_{i=1}^n A_i = B.
    \]
    \item \textbf{Preservation of Marginals:} For each unit \( i \), the marginal probability of treatment remains unchanged, i.e.,
    \[
    \Pr(A_i = 1) = p_i.
    \]
    \item \textbf{Estimation Efficiency:} The joint distribution of the binary assignment vector should be such that when used to estimate treatment effects (e.g., via an inverse probability weighted estimator), the estimator's variance is minimized compared to a baseline method (e.g., independent randomization).
\end{enumerate}

In other words, the problem is to construct a randomized procedure that maps the fractional vector \( p = (p_1, p_2, \dots, p_n) \) (which exactly sums to \( B \)) to a valid binary assignment vector \( A \) that satisfies the above three criteria.

\subsection{Proposed Approach: Swap Rounding}
For the algorithm we use to address this problem and create new treatment assignments, we take inspiration from the swap rounding algorithm described by \citealp{chekuri2010} for \(B\)-uniform matroid polytopes. Swap rounding iteratively selects a pair of fractional entries and transfers as much probability mass as possible between them—using a probabilistic randomized decision—to attempt to drive the entries to 0 or 1 until an integral solution is achieved for all units. We outline this algorithm in psuedo-code in Algorithm \ref{alg:swap_rounding_verbal} (full algorithm described in Appendix).

\begin{algorithm}
\DontPrintSemicolon
\SetAlgoLined
\KwIn{A fractional vector \(p\) with \(\sum_{i=1}^n p_i = B\)}
\KwOut{A binary vector \(A\) with \(\sum_{i=1}^n A_i = B\)}
Set \(p^{(0)} \gets p,\; t\gets0\)\;
\While{\(p^{(t)}\) is not binary}{
  Select two indices \(i,j\) with \(0<p^{(t)}_i,p^{(t)}_j<1\)\;
  \If{\(p^{(t)}_i+p^{(t)}_j\le1\)}{
    \textbf{Case 1:} With probability 
    \(\frac{p^{(t)}_i}{p^{(t)}_i+p^{(t)}_j}\), 
    set \(p^{(t+1)}_i \gets p^{(t)}_i+p^{(t)}_j\) and \(p^{(t+1)}_j \gets 0\); 
    otherwise, set \(p^{(t+1)}_i \gets 0\) and \(p^{(t+1)}_j \gets p^{(t)}_i+p^{(t)}_j\)\;
  }
  \Else{
    \textbf{Case 2:} With probability 
    \(\frac{1-p^{(t)}_i}{2-p^{(t)}_i-p^{(t)}_j}\), round \(p^{(t)}_i\) up to 1 (i.e., set \(p^{(t+1)}_i \gets 1\)) 
    and let \(p^{(t+1)}_j \gets p^{(t)}_i+p^{(t)}_j-1\); 
    otherwise, round \(p^{(t)}_j\) up to 1 and let \(p^{(t+1)}_i \gets p^{(t)}_i+p^{(t)}_j-1\)\;
  }
  For all other indices \(k\), set \(p^{(t+1)}_k \gets p^{(t)}_k\)\;
  Increment \(t\)\;
}
\Return \(A\gets p^{(t)}\)\;
\caption{Concise Swap Rounding Pseudocode with Explicit Probabilities}
\label{alg:swap_rounding_verbal}
\end{algorithm}

The key features of the swap rounding algorithm are as follows:

\begin{itemize}
    \item \textbf{Exact Feasibility:} Starting with the fractional vector \( p \) (with \( \sum_{i=1}^n p_i = B \)), the algorithm iteratively “swaps” probability mass between pairs of units. In each step, it carefully transfers probability so that at the end, every \( p_i \) becomes either 0 or 1, ensuring that exactly \( B \) units are treated.
    \item \textbf{Marginal Preservation:} The swapping is performed in a manner that guarantees the marginal probability for each unit remains equal to its original value \( p_i \). Thus, even after rounding, we have
    \[
    \Pr(A_i = 1) = p_i \quad \text{for all } i.
    \]
    \item \textbf{Negative Correlation:} By coupling the rounding decisions where one probability is increased while the other is decreased, swap rounding induces negative correlations between the binary $A$. We will formally show below that this reduces the variance of treatment effect estimators compared to independent randomization.
\end{itemize}

In our setting, the use of swap rounding is particularly appealing because it simultaneously satisfies the hard resource constraint and preserves the designed assignment probabilities. This dual property is essential for both fair allocation in real-world interventions and for achieving efficient estimation in statistical analyses.

\subsection{Notations and Assumptions} \label{sec:assum}
 Each unit has potential outcomes $Y_i(1)$ and $Y_i(0)$ under treatment and non-treatment respectively. We consider two possible sets of distributional assumptions for the units. We start with the design-based setting where potential outcomes are viewed as fixed. A subset of our key results hold in this setting, e.g., Proposition \ref{prop:unbiased} guaranteeing unbiased estimates and Proposition \ref{Prop:var_red} guaranteeing that swap rounding reduces variance. Then, we impose additional assumptions and move to the superpopulation setting where $Y_i(1),Y_i(0)$ are drawn iid from some distribution $\Pb$. We show that imposing this assumption enables additional results like a central limit theorem and the construction of confidence intervals. In the superpopulation setting, we  optionally consider unit-level covariates. That is, for each unit \(i\), we optionally observe covariates \(V_i\) where $Y_i(0), Y_i(1), V_i$ are jointly sampled iid from some distribution $\Pb$ and the assignment probability is a function of the covariates $p(V_i) = \Pb(A=1| V_i)$. 
 


Given that this is an RCT, we assume ignorability ($Y(1),Y(0) \ind A$)
, and the standard assumptions of consistency ($Y=Y(A)$ whenever $A=a$), and positivity ($p_i, (1-p_i), p(V_i), 1-p(V_i) > 0$). We will also that $Y(0)$ and $Y(1)$ are always nonnegative (this can simply be achieved by scalar translation of $Y$ if it does not hold a-priori). 


\section{Methodology}
\subsection{IPW Estimator}
Our goal in this section is to analyze the statistical properties of the binary treatment assignments produced by our swap rounding procedure. In particular, we focus on the inverse probability weighted (IPW) estimator for the average treatment effect (ATE) when these assignments are used. In later sections, we will further extend our analysis to general linear estimators. Here, we will show that:

\begin{itemize}
    \item The sequence of partial IPW estimators, constructed using the intermediate fractional values from swap rounding, forms a martingale. This property is crucial for establishing that the final estimator is unbiased.
    \item By inducing negative correlations among treatment assignments, swap rounding yields an estimator with a lower variance compared to one based on independent assignment.
\end{itemize}
\subsubsection{Treatment effect estimators}
Our main results compare two ways of drawing the assignment variables and we introduce separate notation for both. First, \(A_i\) is sampled independently with probability \(p_i\). Second, \(A'\) is produced via swap rounding applied to \(\{p_i\}_{i=1}^n\). The corresponding inverse propensity weighted estimators for the average treatment effect (ATE) $\tau$ are
\begin{align}
  \tau_{\text{IPW}} &= \frac{1}{n}\sum_{i=1}^n \frac{A_i\,Y_i(1)}{p_i} - \frac{(1-A_i)\,Y_i(0)}{1-p_i}, \label{eq:tauIPW}\\[1mm]
  \tau_{\text{swap}} &= \frac{1}{n}\sum_{i=1}^n \frac{A'_i\,Y_i(1)}{p_i} - \frac{(1-A'_i)\,Y_i(0)}{1-p_i}. \label{eq:tauSwap}
\end{align}
\subsubsection{Martingale Property of the Estimator}
Let \(p_i^{(0)}\) denote the initial target (fractional) treatment probability for unit \(i\), and let \(p_i^{(t)}\) denote the (partial) probability after the \(t\)th swap rounding step. Furthermore, let $T$ denote the total number of swap-rounding time steps. Note that $T \leq m$ since swap rounding converts at least one of the $m$ entries to an integer at each step and may terminate early if it converts two to integers on a single step (which occurs when $p_i = p_j = 0.5$). Define the IPW-based estimator at step \(t\) as
\begin{align}
x_t &= \frac{1}{n}\sum_{i=1}^n \left(\frac{p_i^{(t)}\,Y_i(1)}{p_i^{(0)}} - \frac{1-p_i^{(t)}}{\,1-p_i^{(0)}}\,Y_i(0)\right).
\end{align}
Because the swap rounding procedure is constructed so that
\begin{align}
\E\Bigl[p_i^{(t)} \,\Big|\, p_i^{(t-1)}\Bigr] &= p_i^{(t-1)},
\end{align}
it follows immediately that
\begin{align}
\E\Bigl[x_t \,\Big|\, x_{t-1}\Bigr] &= x_{t-1}.
\end{align}
Thus, the sequence \(\{x_t\}\) forms a martingale. When \(t = T\) (i.e., after complete rounding), \(p_i^{(T)} = A_i \in \{0,1\}\) and \(x_T\) is the final IPW estimator. This martingale property allows us to  establishing unbiasedness as it implies that $\E[X_T] = E[X_0]$. Formally, let us define the Sample Average Treatment Effect (ATE) as $\tau_{SATE} = \frac{1}{n}\sum_{i = 0}^n Y_i(1) - Y_i(0)$. In the superpopulation setting, we can also define the Population Average Treatment Effect as $\tau_{PATE} = \E[Y(1) - Y(0)]$.

\begin{proposition}[Unbiasedness of IPW Swap Rounding for sample ATE] \label{prop:unbiased} Let $Y_{1:n}(0), Y_{1:n}(1)$ denote the vectors of sample potential outcomes. Conditional on $Y_{1:n}(0), Y_{1:n}(1)$ (and so including in the design-based setting), we have
  \[
\mathbb{E}\left[\tau_{\mathrm{swap}}|Y_{1:n}(0), Y_{1:n}(1)\right] = \tau_{SATE}. 
  \] In the superpopulation setting, we additionally have $\E[\tau_{\mathrm{swap}}] = \tau_{PATE}$, where the expectation is taken over  $A$ and the $Y_{1:n}(0), Y_{1:n}(1)$. 
\end{proposition}
Intuitively, this result holds because swap rounding guarantees that the marginal probability for any individual to be treated is identical as if the assignments were sampled independently, and so the result follows from linearity of expectation. Next, we leverage negative correlation between assignments to show that $\tau_{\text{swap}}$ can only have lower variance than $\tau_{\text{IPW}}$.
\subsubsection{Variance Decomposition}
Define the individual summand for each unit appear in the IPW estimator as
\begin{align}
X_i &= \frac{A'_i\,Y_i(1)}{p_i^{(0)}} - \frac{(1-A'_i)\,Y_i(0)}{\,1-p_i^{(0)}}.
\end{align}
The variance of the final estimator \(x_T\) can be written as
\begin{align}
\mathbb{V}\bigl(x_T\bigr) &= \frac{1}{n^2}\Biggl[\sum_{i=1}^n \mathbb{V}\bigl(X_i\bigr) + 2 \sum_{i,j \in S} \cov\bigl(X_i, X_j\bigr)\Biggr],
\label{eq:6}
\end{align}
where $S$ is the set of $T$ pairs of indices where the swap rounding algorithm made swaps. 
The key advantage of swap rounding is that it introduces negative correlations between the \(A'_i\)’s. In particular, if we denote
\begin{align}
\rho_{ij} &= \cov\bigl(A'_i, A'_j\bigr),
\end{align}
then one can show that
\begin{align}
\rho_{ij} &=
\begin{cases}
-\,p_i^{(0)}\,p_j^{(0)}, & \text{if } p_i^{(0)}+p_j^{(0)} \le 1, \\[1mm]
-\,\bigl(1-p_i^{(0)}\bigr)\bigl(1-p_j^{(0)}\bigr), & \text{if } p_i^{(0)}+p_j^{(0)} > 1.
\end{cases}
\end{align}
This negative covariance (which does not occur with independent sampling) decreases the overall variance of \(x_T\) relative to the estimator based on independent assignments.
\subsubsection{Standard Estimator Variance}
From the variance of the IPW estimator (derivation in appendix), we have $\mathbb{V}(X_i|Y_i(0), Y_i(1))$ as 
\begin{align}
    \frac{Y_i(1)^2}{p_i^0} + \frac{Y_i(0)^2}{1-p_i^0} - [Y_i(1) - Y_i(0)]^2
\end{align}
\subsubsection{Covariance Analysis}
Utilizing the fact that the swap rounding procedure preserves the marginals but induces a negative covariance between the two treatment assignments involved in the swap leads (with omitted proof in the Appendix) to the final covariance expression: $\mathbb{\cov}(X_i,X_j|Y_{i,j}(0), Y_{i,j}(1)) = $
\begin{align}
\rho_{ij} \left(\frac{Y_i(1)}{p_i^{(0)}}
+\frac{Y_i(0)}{1-p_i^{(0)}}\right)
\left(\frac{Y_j(1)}{p_j^{(0)}}
+\frac{Y_j(0)}{1-p_j^{(0)}}\right)
\end{align}

Because of our non-negativity assumption on outcome values and the fact that $\rho_{ij}$ is negative, we see that $\mathbb{\cov}(X_i,X_j)$ is always negative as $\rho_{ij}$ remains negative. Thus, by “coupling” the rounding of different units, our procedure further reduces variance relative to independent assignment.  As 4.4 details the standard variance of the IPW variance, the fact that $\mathbb{\cov}(X_i,X_j)$ is always negative means that we have Proposition \ref{Prop:var_red}.

\begin{proposition}[Variance Reduction]
  Under the swap rounding procedure, the negative dependence between the assignment indicators ensures that for any value of the potential outcomes $Y_{i:n}(0), Y_{i:n}(1)$
  \[
  \mathbb{V}\left(\tau_{\mathrm{swap}}|Y_{i:n}(0), Y_{i:n}(1)\right) \leq \mathbb{V}\left(\tau_{\mathrm{IPW}}|Y_{i:n}(0), Y_{i:n}(1)\right). 
  \]  If there exists at least one swapped pair $i,j$ where $Y_i(0), Y_i(1), Y_j(0), Y_j(1)$ are not all simultaneously zero, the inequality is strict. In the superpopulation setting, we also have unconditionally that $\mathbb{V}\left(\tau_{\mathrm{swap}}\right) \leq \mathbb{V}\left(\tau_{\mathrm{IPW}}\right)$.
  \label{Prop:var_red}
\end{proposition}
This indicates that precision of inference will never decrease if we draw the assignments using swap rounding.

\subsubsection{Inference}
We now turn to providing inferential guarantees, including asymptotically valid confidence intervals, for the swap rounding estimator. Our first task is to construct a consistent estimator for $\mathbb{V}\left(\tau_{\mathrm{swap}}\right)$.  At this stage, all of our results will require the superpopulation assumption that units are sampled iid.  The challenge is that terms of the form $\left[Y_i(1) - Y_i(0)\right]^2$ appear in the variance of the estimator conditional on the potential outcomes, but both potential outcomes are never observed simultaneously for the same unit. In the superpopulation setting, the unconditional variance of the estimator instead contains terms of the form $\E[Y(1) - Y(0)]^2$ that can be proxied with $\tau_{\mathrm{swap}}$ itself. For all other terms, appearing in $\mathbb{V}\left(\tau_{\mathrm{swap}}\right)$, we use inverse-propensity weighted estimates. Let $Y_i = Y_i(A'_i)$ denote the observed outcome for unit $i$. This yields the final variance estimator:

\begin{align}    
    \hat{\sigma} &=
    \frac{1}{n^2} \bigg [   \sum_{i=1}^n    \bigl ( \frac{A'_i (Y_i)^2}{(p_i^0)^2} + \frac{(1-A'_i)(Y_i)^2}{(1-p_i^0)^2}\bigr) -  n\tau_{\mathrm{swap}}^2  
    \\
    &+ 2 \sum_{i,j \in S} \rho_{ij} 
\left(\frac{A'_iY_i}{p_i^{(0)}}
+\frac{(1-A'_i)Y_i}{1-p_i^{(0)}}\right) \\
&\left(\frac{A'_jY_j}{p_j^{(0)}}
+\frac{(1-A'_j)Y_j}{1-p_j^{(0)}}\right)    \bigg]    
\end{align}

As the assignment probabilities $p_i$ are known exactly, we establish that:
\begin{proposition}[Unbiasedness and Consistency of IPW Swap Rounding Variance Estimator] $\hat{\sigma}$ is a consistent and unbiased Estimator of $\mathbb{V}[\tau_\mathrm{swap}]$ under the superpopulation assumption.
  \label{prop:3}
\end{proposition}
Finally, we use a martingale central limit theorem to show that $\tau_\mathrm{swap}$ is asymptotically normal. Combined with Proposition \ref{prop:3}, this enables the construction of valid confidence intervals.
\begin{theorem}[Asymptotic Normality and Valid Confidence Intervals]
Given the superpopulation assumption and under the further assumptions that:,
\begin{itemize}
    \item \textbf{Bounded Moments:} \(Y_{i}(1)\) and \(Y_{i}(0)\) have uniformly bounded second moments by a constant \(M < \infty\).
    \item \textbf{Lower Bound on Quadratic Variation:} There exists a constant \(c > 0\) such that for all sufficiently large \(n\),
    \[
        \bigl\langle x \bigr\rangle_{T}
        \;=\;
        \frac{1}{n}\sum_{t=1}^{T}\mathbb{E}\bigl[\Delta_{t}^{2}\mid x_{t-1}\bigr]
        \;\ge\; c\,.
    \]
\end{itemize}
Them, the IPW estimator obtained via swap rounding, \(\tau_{\mathrm{swap}}\), is asymptotically normal. That is,
\[
\sqrt{n}\left(\tau_{\mathrm{swap}} - \tau\right) \xrightarrow{d} N(0, \sigma^2),
\]
where \(\sigma^2\) is the asymptotic variance given by the expression in Equation \ref{eq:6}. Consequently, we have for \(\widehat{\sigma}^2\), the \(1-\alpha\) confidence interval
\[
\tau_{\mathrm{swap}} \pm z_{1-\alpha/2}\frac{\widehat{\sigma}}{\sqrt{n}}
\]
has asymptotically nominal coverage.
\end{theorem}

\subsection{General Properties of Linear Estimators Under Swap Rounding}
In fact, these properties extend beyond the IPW estimator as we show that a key property of swap rounding is that any estimator that is linear in the treatment assignments is automatically unbiased and has lower variance relative to the case of independent assignments. To state this formally, we have Theorem \ref{thr:1} (omitted proof in Appendix).

\begin{theorem}[Unbiasedness and Variance Reduction for Linear Estimators]
Let 
\begin{align}
  \psi = \sum_{i=1}^n a_i A_i + b
\end{align}
be an estimator that is linear in the treatment assignments \(A_i\) for $a_i \geq 0$. Define $\psi_{swap}$ and $\psi_{IPW}$ as the versions of this estimator where the $A$ are sampled from swap rounding and independently, respectively. Treating $a_i$ and $b$ as fixed and analyzing the estimators conditional on them,
\begin{align}
  \E[\psi_{swap}] &= \E[\psi_{IPW}] = \sum_{i=1}^n a_i\,p_i + b, \\
  \mathbb{V}(\psi_{swap}) &= \sum_{i=1}^n a_i^2\,\mathbb{V}(A'_i) + 2 \sum_{i,j \in S} a_i\,a_j\,\cov(A'_i,A'_j)
  \nonumber \\
  &\le \sum_{i=1}^n a_i^2\,\mathbb{V}(A_i) = \mathbb{V}(\psi_{IPW}),
\end{align}
i.e., the variance is lower than (or equal to) what would be achieved under independent assignment. If there is any pair where $a_i, a_j \neq 0$, the inequality is strict.
\label{thr:1}
\end{theorem}

\section{Covariate-Ordered Swap Rounding} \label{sec:5}

In addition to preserving marginal treatment probabilities and inducing negative correlations through swap rounding, it is beneficial to pair units with similar covariates to further reduce variance. The intuition is that when units are similar in their covariate profiles, they are likely to have similar baseline outcomes. Thus, any difference induced by the randomized rounding is less variable. This idea parallels matching methods in causal inference, where pairing or matching treated and control units based on covariate similarity improves the precision of effect estimates \citep{stuart2010matching}. We will introduce an extension of our swap-rounding methodology that leverages this intuition. 
 
 We first order the \(n\) units by their covariate vectors \(V_i\) using a distance metric 
\[
d(V_i,V_j)=\|V_i-V_j\|,
\]
and solve
\begin{align}
\min_{\pi\in\mathcal{P}(n)}\; \sum_{k=2}^{n} 
d\Bigl(V_{\pi(k)},V_{\pi(k-1)}\Bigr),
\label{eq:opt_ordering}
\end{align}
where \(\mathcal{P}(n)\) denotes the set of all permutations of \(\{1,\dots,n\}\). (This is analogous to a TSP, which can be approximately solved via heuristics \cite{johnson1997local,kirkpatrick1983optimization}.) Swap rounding is then applied only to adjacent pairs in the ordering, ensuring that negative correlations are induced primarily between similar units.

Next, we layout the assumptions needed to ensure covariate ordering swapping reduces variance. While these assumptions are restrictive, this should be seen a stylized model to get a sense of ideal conditions that would exist for covariate ordering to be beneficial, and not something we expect to always hold. Essentially, we want to show that this technique works better for conditions when outcomes are closer for similar covariates than for dissimilar covariates.

Assume the potential outcomes follow
\begin{align}
Y_i(0) &= f(V_i) + \epsilon_{i0}, \label{eq:yi0}\\[1mm]
Y_i(1) &= f(V_i) + \tau(V_i) + \epsilon_{i1}, \label{eq:yi1}
\end{align}
with \(f\) and \(\tau\) Lipschitz (with constants \(L_f\) and \(L_\tau\), respectively), $\epsilon$ as random mean-zero error terms that are independent of the $V$,  and that the target treatment probabilities 
\(p_i^{(0)}\) are Lipschitz in \(V_i\) (with constant \(L_p\)). 

Let \(A'_i\) be the final assignment from swap rounding and 
\[
\mathrm{Cov}(A'_i,A'_j)=\rho_{ij}<0 \quad (i\neq j).
\]
The unit-level IPW estimator is
\[
X_i=\frac{A'_i\,Y_i(1)}{p_i^{(0)}}-\frac{(1-A'_i)\,Y_i(0)}{1-p_i^{(0)}}.
\]
Define the effective weight by
\begin{align}
M_i &\triangleq \frac{\E[Y_i(1)]}{p_i^{(0)}} + \frac{\E[Y_i(0)]}{1-p_i^{(0)}} 
\nonumber\\[1mm]
&\;=\frac{f(V_i)+\tau(V_i)}{p_i^{(0)}}+\frac{f(V_i)}{1-p_i^{(0)}}.
\end{align}
Then, it can be shown that
\[
\mathrm{Cov}(X_i,X_j)=\rho_{ij}\,M_i\,M_j.
\]

To exploit covariate similarity, we assume that \(M(\cdot)\) is \emph{bi-Lipschitz}; i.e., there exist constants 
\(\ell_M,L_M>0\) such that for all \(V_i,V_j\)
\begin{align}
\ell_M\,\|V_i-V_j\| \le |M_i-M_j| \le L_M\,\|V_i-V_j\|.
\label{eq:biLip}
\end{align}
Let \(S_{\mathrm{opt}}\) denote the set of pairs of neighboring units that are swapped under the covariate-optimized pairing and \(S_{\mathrm{rand}}\) be a fixed alternative ordering. We will compare the variance of treatment effect estimates for the swap rounding estimator under \(S_{\mathrm{opt}}\) to \(S_{\mathrm{rand}}\). This will require formalizing the idea that the optimization algorithm succeeds in finding an ordering where neighboring units are all significantly closer than the average under \(S_{\mathrm{rand}}\). First, for \(S_{\mathrm{opt}}\), suppose that
\[
\|V_i-V_j\| \le \delta,
\]
for all $i,j$ for some \(\delta>0\). Hence, 
\begin{align}
|M_i-M_j| \le L_M\,\delta \triangleq \Delta_{\mathrm{opt}}.
\end{align}
In contrast, let the  expected squared covariate distance under \(S_{\mathrm{rand}}\) be
\begin{align}
\mathbb{E}_{(i,j)\sim S_{\mathrm{rand}}}\Bigl[\|V_i-V_j\|^2\Bigr] = c^2,
\label{eq:randSq}
\end{align}
for some \(c>0\). Now, assume that \(c>\delta\), i.e., that the optimized ordering places all neighboring units closer than the average under . Then by the lower bound in \eqref{eq:biLip} and Jensen's inequality,
\begin{align}
\Delta^2_{\mathrm{rand}} &\triangleq \mathbb{E}_{(i,j)\sim S_{\mathrm{rand}}}\Bigl[(M_i-M_j)^2\Bigr]
\ge \ell_M^2\, c^2, \nonumber\\[1mm]
\Delta_{\mathrm{rand}} &\ge \ell_M\, c.
\end{align}
Thus, if
\begin{align}
L_M\,\delta < \ell_M\, c,
\label{eq:pairingCondition}
\end{align}
then \(\Delta_{\mathrm{opt}} < \Delta_{\mathrm{rand}}\). Since
\[
M_iM_j=\left(\frac{M_i+M_j}{2}\right)^2-\left(\frac{M_i-M_j}{2}\right)^2,
\]
a smaller \(|M_i-M_j|\) (for a fixed average) leads to a larger \(M_iM_j\). Consequently, the negative covariance 
\(\rho_{ij}\,M_iM_j\) is stronger under covariate-ordered pairing, reducing the overall variance of the IPW estimator.

\begin{proposition}[Variance Reduction via Covariate-Ordered Pairing]
Assume that units are paired via covariate-ordered swap rounding so that for each pair \((i,j)\) in the optimal pairing 
\(S_{\mathrm{opt}}\) we have 
\[
\|V_i-V_j\| \le \delta,
\]
and that the effective weight function \(M(\cdot)\) satisfies \eqref{eq:biLip}. Further, suppose an arbitrary pairing \(S_{\mathrm{rand}}\) satisfies 
\begin{align}
\mathbb{E}_{(i,j)\sim S_{\mathrm{rand}}}\Bigl[\|V_i-V_j\|^2\Bigr] \ge c^2,
\label{eq:randBoundFinal}
\end{align}
with $L_M\,\delta < \ell_M\, c$ and that condition \eqref{eq:pairingCondition} holds. Then,
\[
\mathbb{V}\bigl(\tau_{\mathrm{IPW}}\bigr)_{S_{\mathrm{opt}}} <
\mathbb{V}\bigl(\tau_{\mathrm{IPW}}\bigr)_{S_{\mathrm{rand}}}.
\]
\end{proposition}

\section{Experiments}
\subsection{General Setup} \label{sec: general}
For our experiments, we generate consider 3 different settings. Firstly, a fully synthetic setting where we generate potential outcomes, assignment probabilities and covariates through simulation. We then consider a semi-synthetic setting involving 2 real datasets.
\subsection{Baselines} \label{sec:baselines}
Below, we briefly describe the baseline methods we compare our swap-rounding IPW estimator and Covariate-Matched Swap-Rounding IPW Estimator to.

\begin{enumerate}
  \item \textbf{Standard IPW Estimator}  
    We use the standard construct of the Horvitz–Thompson IPW estimator
  
  \item \textbf{Budget Limited Bernoulli Assignment:}  
    A conditional assignment procedure that repeatedly draws independent Bernoulli assignments until the draw satisfies \(\sum_i A_i = B\), akin to conditional Poisson sampling \cite{hajek1964,Grafstrm2005ComparisonsOM}.
    
  \item \textbf{Re-randomization with Effective Propensity Adjustment:}  
    Multiple candidate assignments are drawn from the Bernoulli model, and one is selected based on a Mahalanobis distance criterion that enforces covariate balance (see \cite{morgan2012rerandomization}). An effective propensity is then computed using the methodology described in \cite{branson2019maximally} for max sampling for optimal selection and used to adjust the IPW estimator to un-bias the estimates.

    \item \textbf{Random Assignment Based on Total Probabilities:}  
    Exactly \(B\) units are selected uniformly at random (without replacement) for treatment. This corresponds to simple random sampling \cite{cochran1977sampling}. As such, this method does not preserve target probability values $p^{0}$.
    
  \item \textbf{Self-Normalized IPW:}  
    With independent Bernoulli assignments, the self-normalized IPW estimator is computed as
    \[
    \hat{\tau} = \frac{\sum_i A_i Y_i/p_i}{\sum_i A_i/p_i} - \frac{\sum_i (1-A_i)Y_i/(1-p_i)}{\sum_i (1-A_i)/(1-p_i)},
    \]
 This method was designed to reduce the impact of extreme weights and to  reduced variance (which it tends to do extremely well) as compared to standard IPW estimators \cite{liu2016inverse}. However, in finite samples this estimator is not unbiased. \citep{datta_polson_inverse}. While this method and Random Assignment are biased, we include them to show the variance we hope to empirically achieve.
\end{enumerate}
\subsection{Synthetic Experiments}
\subsubsection{Synthetic Experiment Setup} For each sample size $n$ (we use various sample sizes), we generate 100 unique scenarios with covariates \(V \in \mathbb{R}^{n\times3}\) (with entries drawn from a standard normal), and outcomes given by
\[
Y_i(0) = \beta_0 + V_i^\top\beta, \quad Y_i(1) = Y_i(0) + \tau_{\text{true}} + \epsilon_i,
\]
with \(\tau_{\text{true}}=2\) and \(\epsilon_i\sim\mathcal{N}(0,1)\). The distribution of the initial probabilities $p^0$ is computed via the methodology described in section \ref{sec: general}.

In this setting, we consider 3 methodologies for assigning target treatment probabilities $p^{(0)}$.
\begin{enumerate}
    \item \textbf{Uniform Distribution:} $p^{(0)} \sim$ Uniform (0.01,0.99) 
    \item \textbf{Gaussian Distribution:} $p^{(0)} \sim \mathcal{N}(.5,.25)$  with probabilities clipped between (0.01,0.99)
    \item \textbf{Covariate Based Distribution:}  \(p_0\) is computed via a logistic model of $V$ (clipping probabilities between .01 and .99)
\end{enumerate}
 Each of these scenarios is then replicated 100 times. All code is provided is supplementary materials.
\subsubsection{Experiments}
\begin{figure}[h!] 
    \centering
\includegraphics[scale=.25]{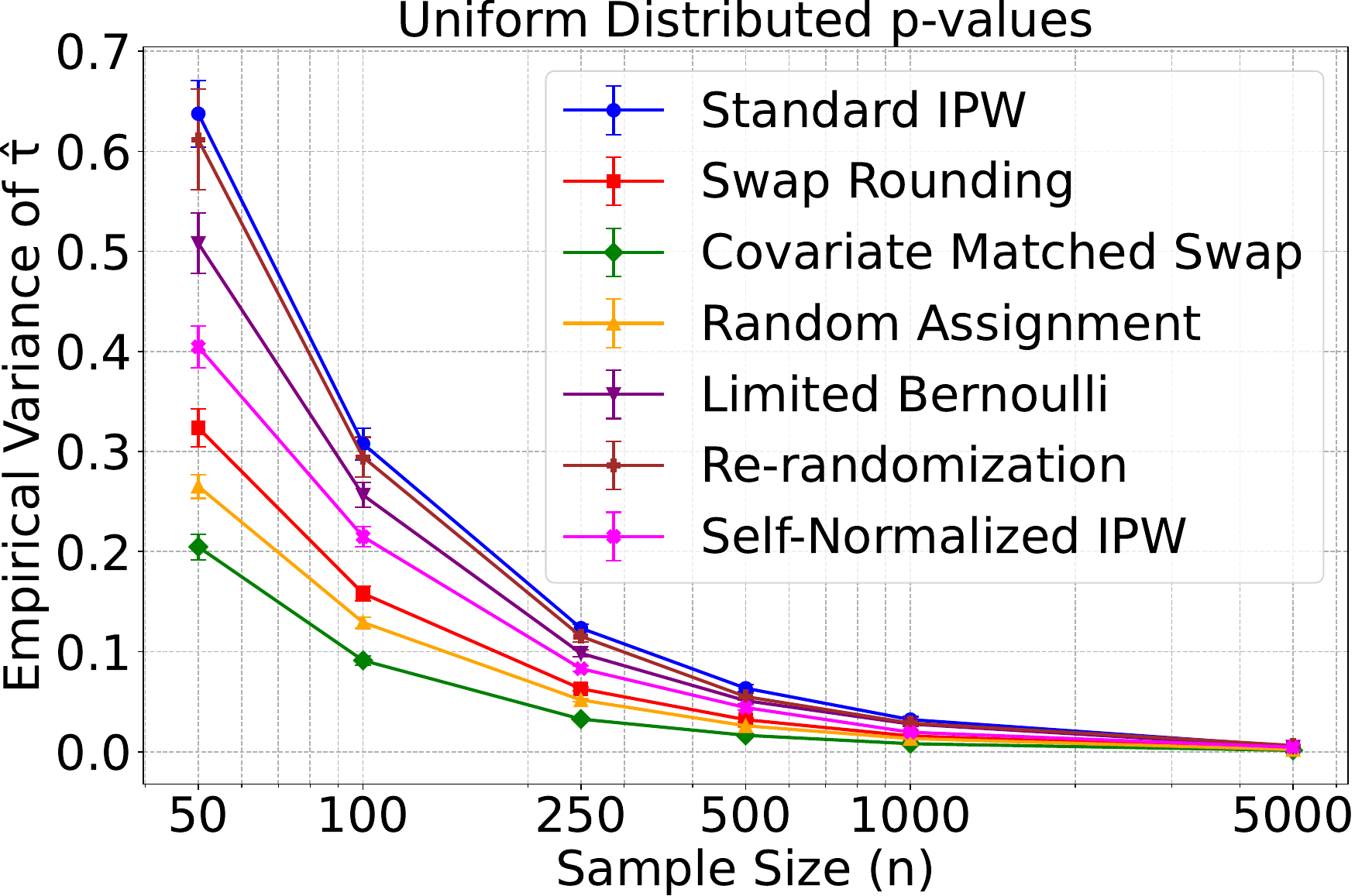}
    \caption{Synthetic Experiment with $p^{(0)} \sim$ Uniform (0.01,0.99). Mean empirical variance with 95 $\%$ CI.   }
    \label{fig:1}
\end{figure}

\begin{figure}[h!] 
    \centering
\includegraphics[scale=.25]{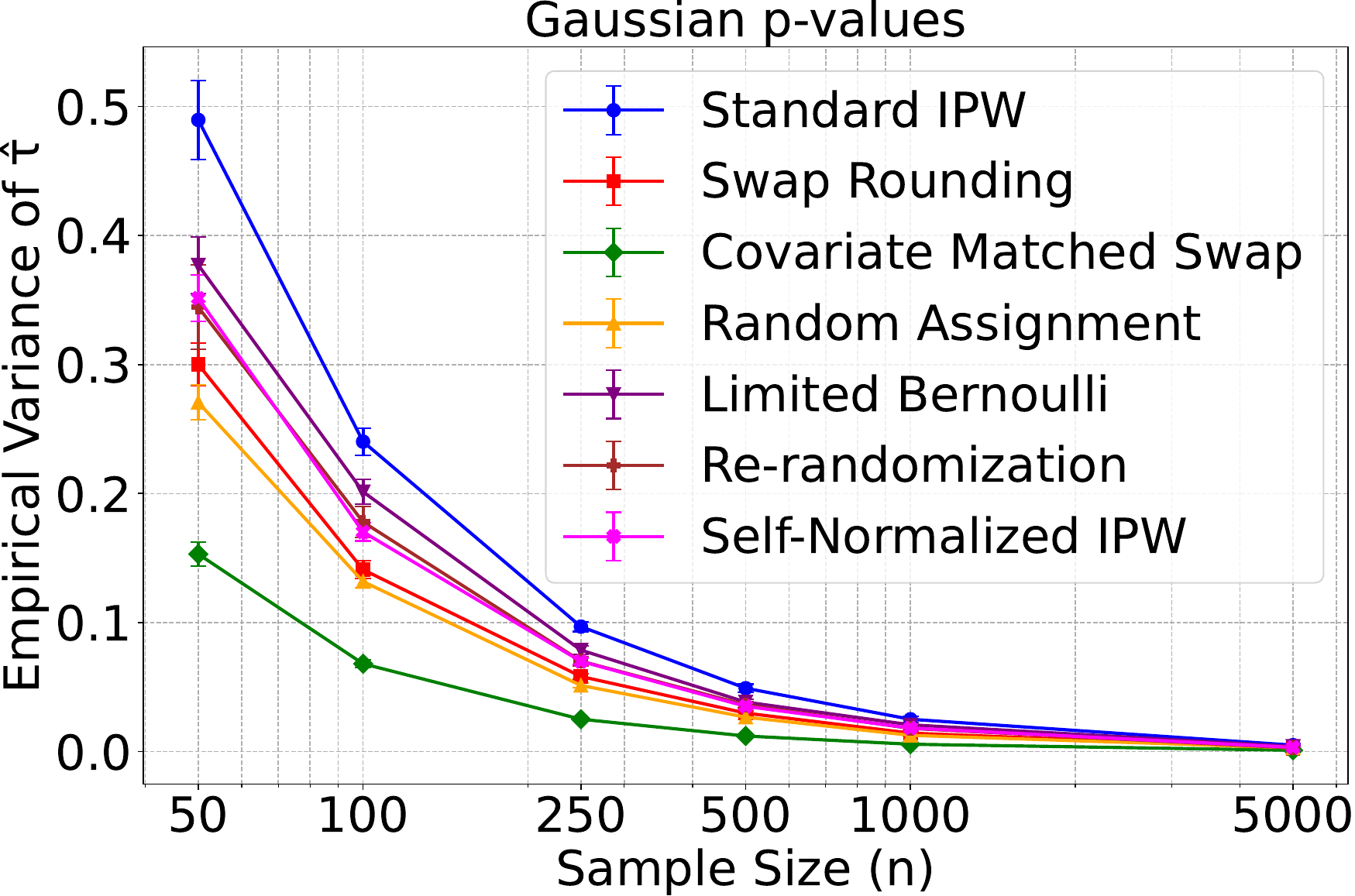}
    \caption{Synthetic Experiment with $p^{(0)} \sim \mathcal{N}(.5,.25)$  }
    \label{fig:2}
\end{figure}

\begin{figure}[h!] 
    \centering
\includegraphics[scale=.25]{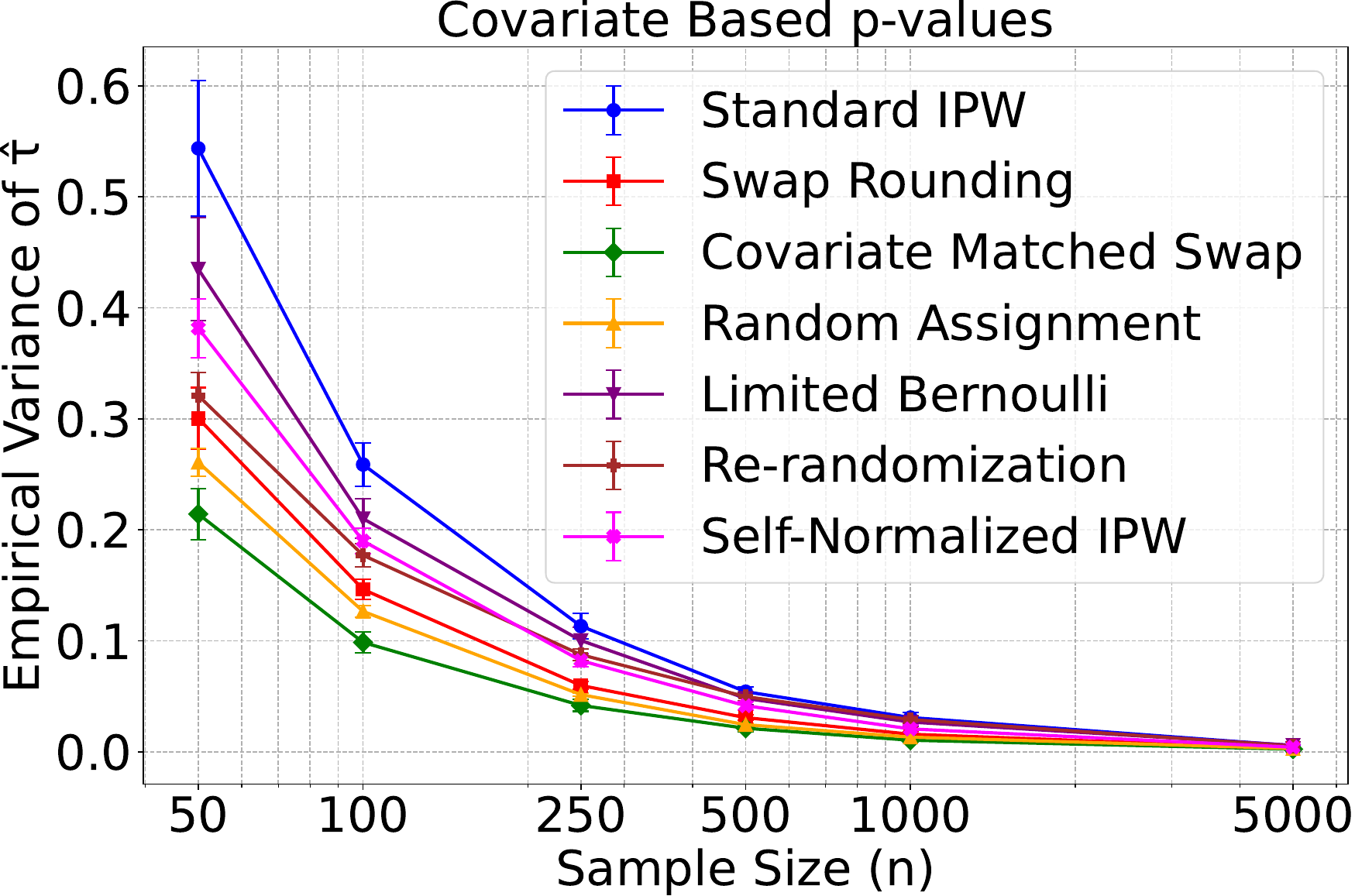}
        \caption{Synthetic Experiment where \(p_0\) is computed via a logistic model of the covariates $V$  }
    \label{fig:3}
\end{figure}

We conduct 3 separate experiments, each one different by the methodology of how the target probabilities $p^{(0)}$ are distributed. In all 3 experiments (Uniform, Gaussian and Covariate-Based Distribution seen in Figures \ref{fig:1}, \ref{fig:2}, \ref{fig:3}), we notice a general trend. Covariate Based Swap Rounding performs the best (minimum variance) by a considerable margin, followed by Random Assignment, then standard Swap Rounding, Self-Normalized IPW and then other methods. This result is especially significant because our methodology outperforms estimators like Self-Normalized IPW and Random Assignment that were designed to minimize variance at the cost of bias. We also notice several other interesting trends such as the standard IPW estimator performing worst in all experiments. At large sample sizes, all estimators converge to negligible variance.
 \subsection{Semi-Synthetic Data}
 For the type of data that is needed to test these methods, we need access to the true propensity scores underlying trial selection as well as access to both potential outcomes for each individual. This obviously is not possible so we have to sacrifice data quality in some places and make it up with simulation. That being said, we proceed with two data sources that provide different levels of information.
 \subsubsection{IDHP Dataset}
\begin{figure}[h!] 
    \centering
\includegraphics[scale=.25]{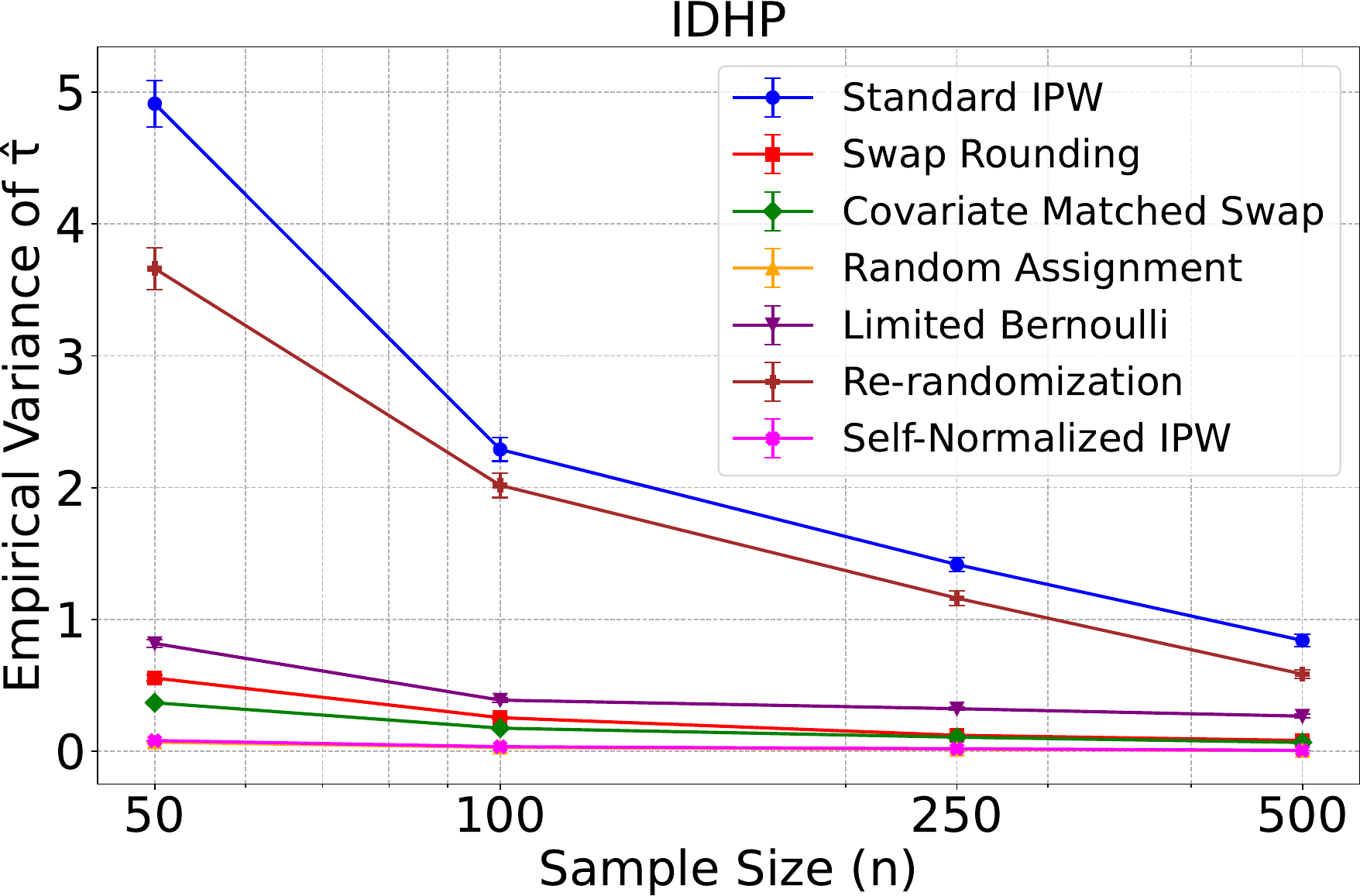}
    \caption{Semi-Synthetic Experiment IDHP dataset }
    \label{fig:4}
\end{figure}

 Our first data source is the IDHP Dataset. \citep{ihdp2024} The dataset (747 individuals) is based off an RCT that was conducted between 1985-1988 where they tested the impact of early interventions in reducing developmental problems associated with low weight and premature infants where the outcome was a continuous measure of cognitive development. We used simulated counterfactual outcomes for each individual as well propensity scores estimated via logistic regression. Our results are shown in Figure \ref{fig:4} (the Random Assignment line is hidden underneath the Self-Normalized IPW line). In this setting again our methodology remains competitive with the biased methodologies while performing superior to those methods who are unbiased. 
 \subsubsection{Housing Dataset}
\begin{figure}[h!] 
    \centering
\includegraphics[scale=.25]{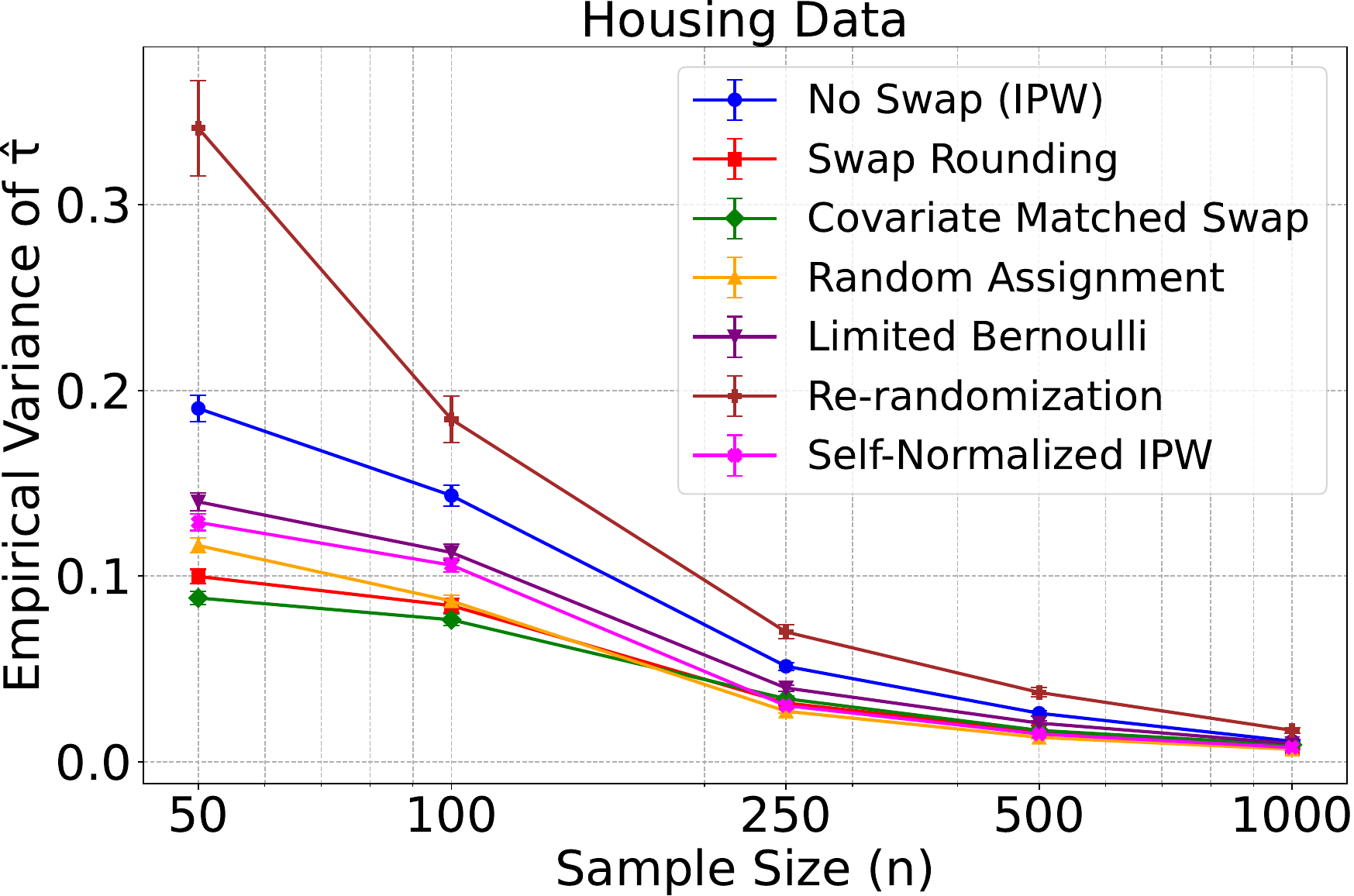}
    \caption{Semi-Synthetic Experiment Housing Dataset }
    \label{fig:5}
\end{figure}

 In this setting, we use data provided by the Allegheny County Department of Human Services. The dataset (3163 individuals) was introduced by \citet{wilder2024learningtreatmenteffectstreating} and contains individuals eligible for public housing assistance and contains assignment probabilities as well as simulates the effect of an intervention on the binary outcome whether they will have at least 4 emergency room visits in the upcoming year.
 The hypothetical intervention’s effect is characterized by $\tau(X) = .1 \cdot P(Y(0)=1 \mid X$)
where 0.1 quantifies the proportional decrease in the occurrence of an adverse outcomes resulting from the treatment, and the probability \(P(Y(0)=1 \mid X\bigr)\) is estimated using a random forest model. Our results are shown in Figure \ref{fig:5}. In this experiment, regular swap rounding performs the best. 
\section{Discussion} 
We theoretically show that incorporating swap rounding, due the negative correlations it induces, leads to an unbiased and variance-reduced IPW estimator that can enforce strict budget constraints. Moreover, by incorporating covariate-ordered pairing, our approach further refines variance reduction, particularly in settings where units share similar characteristics. These findings highlight the potential of our method for designing efficient and fair experiments in resource-constrained environments.

\section{Acknowledgments}
This project was supported by the AI2050 program at Schmidt Sciences (Grant G-22-64474) and the AI Research Institutes Program funded by the National Science Foundation under AI Institute for Societal Decision Making (AI-SDM), Award No. 2229881. We would also like to acknowledge Naveen Raman, Nima Jadali and Albert Ting for their contributions in editing and advising on the paper. 
\newpage
\bibliography{uai2025-template.bib}

\begin{thebibliography}{24}
\providecommand{\natexlab}[1]{#1}
\providecommand{\url}[1]{\texttt{#1}}
\expandafter\ifx\csname urlstyle\endcsname\relax
  \providecommand{\doi}[1]{doi: #1}\else
  \providecommand{\doi}{doi: \begingroup \urlstyle{rm}\Url}\fi

\bibitem[Ageev and Sviridenko(2004)]{Ageev2004}
A.~A. Ageev and M.~Sviridenko.
\newblock Approximation algorithms for maximization of submodular functions.
\newblock In \emph{Proceedings of the 35th Annual ACM Symposium on Theory of Computing (STOC)}, pages 793--802, 2004.

\bibitem[Aires(2000)]{AIRES2000133}
Nibia Aires.
\newblock Comparisons between conditional poisson sampling and pareto πps sampling designs.
\newblock \emph{Journal of Statistical Planning and Inference}, 88\penalty0 (1):\penalty0 133--147, 2000.
\newblock ISSN 0378-3758.
\newblock \doi{https://doi.org/10.1016/S0378-3758(99)00205-0}.
\newblock URL \url{https://www.sciencedirect.com/science/article/pii/S0378375899002050}.

\bibitem[Athey and Imbens(2017)]{Athey2017}
S.~Athey and G.~W. Imbens.
\newblock The state of applied econometrics: Causality and policy evaluation.
\newblock \emph{Journal of Economic Perspectives}, 31\penalty0 (2):\penalty0 3--32, 2017.

\bibitem[Branson and Miao(2019)]{branson2019maximally}
R~Branson and Q~Miao.
\newblock Maximally balanced sampling in randomized experiments.
\newblock \emph{Journal of Statistical Planning and Inference}, 205:\penalty0 1--10, 2019.

\bibitem[Chassang et~al.(2012)Chassang, Padró I~Miquel, and Snowberg]{10.1257/aer.102.4.1279}
Sylvain Chassang, Gerard Padró I~Miquel, and Erik Snowberg.
\newblock Selective trials: A principal-agent approach to randomized controlled experiments.
\newblock \emph{American Economic Review}, 102\penalty0 (4):\penalty0 1279–1309, June 2012.
\newblock \doi{10.1257/aer.102.4.1279}.
\newblock URL \url{https://www.aeaweb.org/articles?id=10.1257/aer.102.4.1279}.

\bibitem[Chekuri and Vondr{\'{a}}k(2009)]{chekuri2010}
Chandra Chekuri and Jan Vondr{\'{a}}k.
\newblock Randomized pipage rounding for matroid polytopes and applications.
\newblock \emph{CoRR}, abs/0909.4348, 2009.
\newblock URL \url{http://arxiv.org/abs/0909.4348}.

\bibitem[Cochran(1977)]{cochran1977sampling}
William~G Cochran.
\newblock \emph{Sampling Techniques}.
\newblock John Wiley \& Sons, 1977.

\bibitem[Datta and Polson(2011)]{datta_polson_inverse}
Jyotishka Datta and Nicholas~G. Polson.
\newblock Inverse probability weighting: from survey sampling to evidence estimation.
\newblock Working paper, Department of Statistics, Virginia Tech \\ and \\ Booth School of Business, University of Chicago, 2011.
\newblock Unpublished manuscript.

\bibitem[Grafstr{\"o}m(2005)]{Grafstrm2005ComparisonsOM}
Anton Grafstr{\"o}m.
\newblock Comparisons of methods for generating conditional poisson samples and sampford samples.
\newblock 2005.
\newblock URL \url{https://api.semanticscholar.org/CorpusID:56873261}.

\bibitem[Gross(2024)]{ihdp2024}
Ruth~T. Gross.
\newblock Infant health and development program (ihdp): Enhancing the outcomes of low birth weight, premature infants in the united states, 1985--1988.
\newblock Inter-university Consortium for Political and Social Research, 2024.
\newblock URL \url{https://doi.org/10.3886/ICPSR09795.v2}.
\newblock Dataset, Version V2, released February 14, 2024.

\bibitem[H{\'a}jek(1964)]{hajek1964}
J~H{\'a}jek.
\newblock Asymptotic theory of rejective sampling with varying probabilities.
\newblock \emph{The Annals of Mathematical Statistics}, pages 1006--1018, 1964.

\bibitem[Imbens and Rubin(2015)]{Imbens2015causal}
G.~W. Imbens and D.~B. Rubin.
\newblock \emph{Causal Inference in Statistics, Social, and Biomedical Sciences}.
\newblock Cambridge University Press, 2015.

\bibitem[Isaki and Fuller(1984)]{Isaki1984}
C.~Isaki and W.~A. Fuller.
\newblock Estimation for finite population sampling.
\newblock \emph{Journal of the American Statistical Association}, 79\penalty0 (387):\penalty0 137--145, 1984.

\bibitem[Johnson and McGeoch(1997)]{johnson1997local}
David~S. Johnson and Laurence~A. McGeoch.
\newblock The traveling salesman problem: A case study in local optimization.
\newblock In \emph{Local Search in Combinatorial Optimization}, pages 215--310. Princeton University Press, 1997.

\bibitem[Kido(2023)]{kido2023incorporatingpreferencestreatmentassignment}
Daido Kido.
\newblock Incorporating preferences into treatment assignment problems, 2023.
\newblock URL \url{https://arxiv.org/abs/2311.08963}.

\bibitem[Kirkpatrick et~al.(1983)Kirkpatrick, Gelatt, and Vecchi]{kirkpatrick1983optimization}
S.~Kirkpatrick, C.~D. Gelatt, and M.~P. Vecchi.
\newblock Optimization by simulated annealing.
\newblock \emph{Science}, 220\penalty0 (4598):\penalty0 671--680, 1983.

\bibitem[Li et~al.(2018)Li, Ding, and Rubin]{Li2018rerandomization}
X.~Li, P.~Ding, and D.~B. Rubin.
\newblock Rerandomization in experiments: A review.
\newblock \emph{Annual Review of Statistics and Its Application}, 5:\penalty0 1--19, 2018.

\bibitem[Liu et~al.(2016)Liu, Hudgens, and Becker-Dreps]{liu2016inverse}
Lan Liu, Michael~G Hudgens, and Sara Becker-Dreps.
\newblock On inverse probability-weighted estimators in the presence of interference.
\newblock \emph{Biometrika}, 103\penalty0 (4):\penalty0 829--842, 2016.
\newblock \doi{10.1093/biomet/asw047}.

\bibitem[Morgan and Rubin(2012)]{morgan2012rerandomization}
D.~L. Morgan and D.~B. Rubin.
\newblock Rerandomization to improve covariate balance in experiments.
\newblock \emph{Journal of the Royal Statistical Society: Series B (Statistical Methodology)}, 74\penalty0 (4):\penalty0 515--532, 2012.

\bibitem[Narita(2021)]{7doi:10.1073/pnas.2008740118}
Yusuke Narita.
\newblock Incorporating ethics and welfare into randomized experiments.
\newblock \emph{Proceedings of the National Academy of Sciences}, 118\penalty0 (1):\penalty0 e2008740118, 2021.
\newblock \doi{10.1073/pnas.2008740118}.
\newblock URL \url{https://www.pnas.org/doi/abs/10.1073/pnas.2008740118}.

\bibitem[Neyman(1934)]{Neyman1934}
J.~Neyman.
\newblock On the problem of the most efficient allocation of units in stratified sampling.
\newblock \emph{Journal of the Royal Statistical Society}, 97:\penalty0 544--557, 1934.

\bibitem[Srinivasan(2001)]{Srinivasan2001}
A.~Srinivasan.
\newblock Distributions on level-sets with applications to approximation algorithms.
\newblock In \emph{Proceedings of the 42nd IEEE Symposium on Foundations of Computer Science (FOCS)}, pages 588--597, 2001.

\bibitem[Stuart(2010)]{stuart2010matching}
Elizabeth~A. Stuart.
\newblock Matching methods for causal inference: A review and a look forward.
\newblock \emph{Statistical Science}, 25\penalty0 (1):\penalty0 1--21, 2010.

\bibitem[Wilder and Welle(2024)]{wilder2024learningtreatmenteffectstreating}
Bryan Wilder and Pim Welle.
\newblock Learning treatment effects while treating those in need, 2024.
\newblock URL \url{https://arxiv.org/abs/2407.07596}.

\end{thebibliography}
\newpage

\onecolumn
\title{Supplementary Material}
\section{Swap Rounding Algorithm}
\begin{algorithm}
\DontPrintSemicolon
\SetAlgoLined
\KwIn{Fractional vector \(p\in[0,1]^n\) with \(\sum_{i=1}^n p_i = B\)}
\KwOut{Binary vector \(A\in\{0,1\}^n\) with \(\sum_{i=1}^n A_i = B\)}
\(t\gets 0,\; p^{(0)}\gets p\)\;
\While{\(p^{(t)}\) is not integral, }{
  Select indices \(i\) and \(j\) such that \(0 < p^{(t)}_i < 1\) and \(0 < p^{(t)}_j < 1\)\;
  \eIf{\(p^{(t)}_i+p^{(t)}_j\le 1\)}{
    Let \(\delta=\min\{p^{(t)}_i,\,p^{(t)}_j\}\)\;
    With probability \(\frac{p^{(t)}_i}{p^{(t)}_i+p^{(t)}_j}\), set 
    \[
      p^{(t+1)}_i = p^{(t)}_i-\delta,\quad p^{(t+1)}_j = p^{(t)}_j+\delta;
    \]
    otherwise set 
    \[
      p^{(t+1)}_i = p^{(t)}_i+\delta,\quad p^{(t+1)}_j = p^{(t)}_j-\delta;
    \]
  }{
    Let \(\delta=\min\{1-p^{(t)}_i,\,1-p^{(t)}_j\}\)\;
    With probability \(\frac{1-p^{(t)}_i}{2-p^{(t)}_i-p^{(t)}_j}\), set 
    \[
      p^{(t+1)}_i = p^{(t)}_i+\delta,\quad p^{(t+1)}_j = p^{(t)}_j-\delta;
    \]
    otherwise set 
    \[
      p^{(t+1)}_i = p^{(t)}_i-\delta,\quad p^{(t+1)}_j = p^{(t)}_j+\delta;
    \]
  }
  For all \(k\notin\{i,j\}\), let \(p^{(t+1)}_k=p^{(t)}_k\)\;
  \(t\gets t+1\)\;
}
\Return \(A=p^{(t)}\)\;
\caption{Simplified Swap Rounding for \(B\)-Uniform Matroid}
\label{algo:swap_rounding}
\end{algorithm}
\newpage
\section{IPW Proofs}
\subsection{IPSW- Swap Rounding}
We want to figure out which subjects to assign treatments to so that we can use an IPSW estimator to estimate $\hat{\tau}$ - the average treatment effect. We start with a vector of probabilities p. This vector then undergoes iterative swap rounding results in A, a vector of binary treatments. The negative correlation aspect of swap rounding allows us to maintain concentration bounds when we don't have a very large number of samples. We observe the outcomes Y(A) and use them in our IPSW estimator for $\hat{\tau}$. We define the following variables:
\\
$p_t$: probabilities after step t of swap rounding \\
$p_0$: target probabilites (pre-rounding) \\
$p_T$: $A_1, ..., A_n$ (binary treatments)
\begin{align*}
    x_t &= \frac{1}{n} \sum_{i } \frac{p_i^t Y_i(1)}{p_i^0} - \frac{(1-p_{i}^t)Y_i(0)}{1-p_{i}^0}  \\
    x_0 &= \frac{1}{n} \sum_i Y(1) - Y(0) \\
    x_T &= \frac{1}{n }\sum_i \frac{A_i Y(1)}{p_i^0} - \frac{ (1-A_{i})Y(0)}{1-p_{i}^0}
\end{align*}

\subsection{Showing Martingale Property}
First, we want to show that
\begin{align*}
    \E[x_t | x_{t-1}] = x_{t-1}
\end{align*}
We have
\begin{align*}
    \E[x_t | x_{t-1}] &= \frac{1}{n} \sum_{i } \frac{\E[p_i^t|p_i^{t-1}] Y_i(1)}{p_i^0} - \frac{(1-\E[p_{i}^t |p_{i}^{t-1} ])Y_i(0)}{1-p_{i}^0} \\
    &= \frac{1}{n} \sum_{i } \frac{p_i^{t-1} Y_i(1)}{p_i^0} - \frac{(1-p_{i}^{t-1}) Y_i(0)}{1-p_{i}^0} \\
    &= x_{t-1}
\end{align*}

\subsection{Variance-Covariance Decomposition}
\begin{align*}
    \mathbb{V}[X_T] = \frac{1}{n^2} \bigl (   \sum_{i=1}^N   \mathbb{V} (\frac{A_i Y_i(1)}{p_i^0} - \frac{ (1-A_{i})Y_i(0)}{1-p_{i}^0}) + 2 \sum_{i,j \in S}\text{Cov}(\frac{A_i Y_i(1)}{p_i^0} - \frac{ (1-A_{i})Y_i(0)}{1-p_{i}^0},\frac{A_j Y_j(1)}{p_j^0} - \frac{ (1-A_{j})Y_j(0)}{1-p_{j}^0})     \bigr )
\end{align*}

where S is the set of swaps in order. For the variance component, we have 

\section*{Single-Unit Variance for Design Based Case}
Define
\[
X_i \;=\; \frac{A_i'\,Y_i(1)}{p_i^0} \;-\; \frac{(1 - A_i')\,Y_i(0)}{1 - p_i^0}.
\]
Then
\begin{align*}
X_i^2 
&= \left(\frac{A_i'\,Y_i(1)}{p_i^0} \;-\; \frac{(1 - A_i')\,Y_i(0)}{1 - p_i^0}\right)^{\!2} \\[6pt]
&= \frac{A_i'\,Y_i(1)^2}{(p_i^0)^2}
 \;+\; \frac{(1 - A_i')\,Y_i(0)^2}{(1 - p_i^0)^2}
 \;-\; 2\,\frac{A_i'\,(1 - A_i')\,Y_i(1)\,Y_i(0)}{p_i^0\,(1 - p_i^0)} \,.
\end{align*}
Taking expectations (using \(\mathbb{E}[A_i'] = p_i^0\) and \(\mathbb{E}[A_i'\,(1 - A_i')] = 0\)) yields
\[
\mathbb{E}[X_i^2]
\;=\;
\frac{Y_i(1)^2}{p_i^0} \;+\; \frac{Y_i(0)^2}{1 - p_i^0}.
\]
Since \(\mathbb{E}[X_i] = Y_i(1) - Y_i(0)\), it follows that
\[
\mathrm{Var}(X_i)
\;=\;
\frac{Y_i(1)^2}{p_i^0} \;+\; \frac{Y_i(0)^2}{1 - p_i^0}
\;-\; \bigl[Y_i(1) - Y_i(0)\bigr]^{2}.
\]

\section*{Super Population}

If \(\bigl(Y_i(0),\,Y_i(1),\,p_i^0\bigr)\) is drawn i.i.d.\ from a larger (hypothetically infinite) population, then taking an outer expectation of the above result gives
\[
\mathbb{V}(X_i)
\;=\;
\frac{\mathbb{E}\bigl[Y_i(1)^2\bigr]}{p_i^0}
\;+\;
\frac{\mathbb{E}\bigl[Y_i(0)^2\bigr]}{1 - p_i^0}
\;-\;
\mathbb{E}\bigl[Y_i(1) - Y_i(0)\bigr]^{2}.
\]

\subsection{Covariance of $X_i$ and $X_j$ in a Design Based Case}
Now, we aim to calculate the covariance - for a pair of units $X_i, X_j$
\[
\text{Cov}(X_i, X_j) = \mathbb{\E}[X_i X_j] - \mathbb{\E}[X_i]\mathbb{\E}[X_j],
\]
where
\[
X_i = \frac{A_i Y_i(1)}{p_i^0} - \frac{(1 - A_i) Y_i(0)}{1 - p_i^0}, \quad
X_j = \frac{A_j Y_j(1)}{p_j^0} - \frac{(1 - A_j) Y_j(0)}{1 - p_j^0}.
\]

\subsection{Expanding $X_i X_j$}
The product $X_i X_j$ expands as:
\begin{align*}
X_i X_j &= \left( \frac{A_i Y_i(1)}{p_i^0} - \frac{(1 - A_i) Y_i(0)}{1 - p_i^0} \right)
\left( \frac{A_j Y_j(1)}{p_j^0} - \frac{(1 - A_j) Y_j(0)}{1 - p_j^0} \right) \\
&= \frac{A_i A_j Y_i(1) Y_j(1)}{p_i^0 p_j^0}
- \frac{A_i (1 - A_j) Y_i(1) Y_j(0)}{p_i^0 (1 - p_j^0)}
- \frac{(1 - A_i) A_j Y_i(0) Y_j(1)}{(1 - p_i^0) p_j^0} \\
&\quad + \frac{(1 - A_i)(1 - A_j) Y_i(0) Y_j(0)}{(1 - p_i^0)(1 - p_j^0)}.
\end{align*}

\subsection{Taking Expectations}
\subsubsection*{First Term: $\mathbb{\E}\left[\frac{A_i A_j Y_i(1) Y_j(1)}{p_i^0 p_j^0}\right]$}
Define $\rho_{ij} = \text{Cov}(A_i, A_j)$. We then have 
\[
\mathbb{\E}[A_i A_j] = p_i^0 p_j^0 + \rho_{ij}.
\]
Thus:
\[
\mathbb{\E}\left[\frac{A_i A_j Y_i(1) Y_j(1)}{p_i^0 p_j^0}\right] =
\frac{(p_i^0 p_j^0 + \rho_{ij}) \cdot Y_i(1)Y_j(1)}{p_i^0 p_j^0}.
\]

\subsubsection*{Second Term: $\mathbb{\E}\left[\frac{A_i (1 - A_j) Y_i(1) Y_j(0)}{p_i^0 (1 - p_j^0)}\right]$}

\[
\mathbb{\E}[A_i (1 - A_j)] = p_i^0 (1 - p_j^0) - \rho_{ij}.
\]
Thus:
\[
\mathbb{\E}\left[\frac{A_i (1 - A_j) Y_i(1) Y_j(0)}{p_i^0 (1 - p_j^0)}\right] =
\frac{(p_i^0 (1 - p_j^0) - \rho_{ij}) \cdot Y_i(1)  Y_j(0)]}{p_i^0 (1 - p_j^0)}.
\]

\subsubsection*{Third Term: $\mathbb{\E}\left[\frac{(1 - A_i) A_j Y_i(0) Y_j(1)}{(1 - p_i^0) p_j^0}\right]$}

\[
\mathbb{\E}[(1 - A_i) A_j] = (1 - p_i^0) p_j^0 - \rho_{ij}.
\]
Thus:
\[
\mathbb{\E}\left[\frac{(1 - A_i) A_j Y_i(0) Y_j(1)}{(1 - p_i^0) p_j^0}\right] =
\frac{((1 - p_i^0) p_j^0 - \rho_{ij}) \cdot Y_i(0)Y_j(1)}{(1 - p_i^0) p_j^0}.
\]

\subsubsection*{Fourth Term: $\mathbb{\E}\left[\frac{(1 - A_i)(1 - A_j) Y_i(0) Y_j(0)}{(1 - p_i^0)(1 - p_j^0)}\right]$}

\[
\mathbb{\E}[(1 - A_i)(1 - A_j)] = (1 - p_i^0)(1 - p_j^0) + \rho_{ij}.
\]
Thus:
\[
\mathbb{\E}\left[\frac{(1 - A_i)(1 - A_j) Y_i(0) Y_j(0)}{(1 - p_i^0)(1 - p_j^0)}\right] =
\frac{((1 - p_i^0)(1 - p_j^0) + \rho_{ij}) \cdot Y_i(0)Y_j(0)}{(1 - p_i^0)(1 - p_j^0)}.
\]

\subsection{Subtracting $\mathbb{\E}[X_i]\mathbb{\E}[X_j]$}
The product $\mathbb{\E}[X_i] \mathbb{\E}[X_j]$ is:
\[
\mathbb{\E}[X_i] \mathbb{\E}[X_j] =
\left( \frac{p_i^0 \cdot Y_i(1)}{p_i^0} - \frac{(1 - p_i^0) \cdot Y_i(0)}{1 - p_i^0} \right)
\cdot \left( \frac{p_j^0 \cdot Y_j(1)}{p_j^0} - \frac{(1 - p_j^0) \cdot Y_j(0)}{1 - p_j^0} \right).
\]
After subtracting from $\mathbb{\E}[X_i X_j]$, we get

\subsection{Final Covariance Expression}
The covariance $\text{Cov}(X_i, X_j)$ is:
\begin{align*}
\text{Cov}(X_i, X_j) &=  \rho_{ij} \cdot \Bigg(
\frac{Y_i(1)Y_j(1)}{p_i^0 p_j^0}
+ \frac{Y_i(1)Y_j(0)}{p_i^0 (1 - p_j^0)} \\
&\quad + \frac{Y_i(0)Y_j(1)}{(1 - p_i^0) p_j^0}
+ \frac{Y_i(0)Y_j(0)}{(1 - p_i^0)(1 - p_j^0)}
\Bigg).
\end{align*}
where
\begin{align*}
    \rho_{ij} &= \text{Cov} (A_i, A_j) =  \E[A_iA_j] - p_i^0p_j^0
\end{align*}
We then have

\[
\mathbb{\E}[A_i A_j] =
\begin{cases}
0 & \text{if } p_i^0 + p_j^0 \leq 1, \\
p_i^0 + p_j^0 - 1 & \text{if } p_i^0 + p_j^0 > 1.
\end{cases}
\]
so

\[
\rho_{ij} =
\begin{cases}
- p_i^0p_j^0 & \text{if } p_i^0 + p_j^0 \leq 1, \\
- (1-p_i^0)(1-p_j^0)& \text{if } p_i^0 + p_j^0 > 1.
\end{cases}
\]
\subsection{Variance Reduction}
As the covariance is guaranteed to be negative due to the negativity of $\rho$, we have a reduction in variance given the non-negativity assumption we make on the outcome. 
\subsection{Super-Population Assumption}
If we employ the super-population assumption, we get 
 covariance $\text{Cov}(X_i, X_j)$:
\begin{align*}
\text{Cov}(X_i, X_j) &=  \rho_{ij} \cdot \Bigg(
\frac{\mathbb{\E}[Y_i(1)] \cdot \mathbb{\E}[Y_j(1)]}{p_i^0 p_j^0}
+ \frac{\mathbb{\E}[Y_i(1)] \cdot \mathbb{\E}[Y_j(0)]}{p_i^0 (1 - p_j^0)} \\
&\quad + \frac{\mathbb{\E}[Y_i(0)] \cdot \mathbb{\E}[Y_j(1)]}{(1 - p_i^0) p_j^0}
+ \frac{\mathbb{\E}[Y_i(0)] \cdot \mathbb{\E}[Y_j(0)]}{(1 - p_i^0)(1 - p_j^0)}
\Bigg).
\end{align*}
\subsection{Variance of $X_T$ Expression}
\begin{align*}
     \mathbb{V}[X_T]  &=
    \frac{1}{n^2} \bigg [   \sum_{i=1}^N    \bigl ( \frac{\E[Y_i(1)^2]}{p_i^0} + \frac{\E[Y_i(0)^2]}{1-p_i^0}\bigr)  - n\mathbb{\E} [Y(1) - Y(0)]^2
    \\
    &+ 2 \sum_{i,j \in S} \rho_{ij} \cdot \Bigl(
\frac{\mathbb{\E}[Y_i(1)] \cdot \mathbb{\E}[Y_j(1)]}{p_i^0 p_j^0}
- \frac{\mathbb{\E}[Y_i(1)] \cdot \mathbb{\E}[Y_j(0)]}{p_i^0 (1 - p_j^0)} \\
&\quad - \frac{\mathbb{\E}[Y_i(0)] \cdot \mathbb{\E}[Y_j(1)]}{(1 - p_i^0) p_j^0}
+ \frac{\mathbb{\E}[Y_i(0)] \cdot \mathbb{\E}[Y_j(0)]}{(1 - p_i^0)(1 - p_j^0)} \Bigr)   \bigg]
    \end{align*}

\subsection{Estimated Variance (IPW)}
\begin{align*}
     \mathbb{V}[X_T]  &=
    \frac{1}{n^2} \bigg [   \sum_{i=1}^N    \bigl ( \frac{A_i (Y_i)^2}{(p_i^0)^2} + \frac{(1-A_i)(Y_i)^2}{(1-p_i^0)^2}\bigr) -  n\hat{\tau}^2  
    \\
    &+ 2 \sum_{i,j \in S} \rho_{ij} \cdot \Bigl(
\frac{A_i A_j Y_i  Y_j}{(p_i^0)^2 (p_j^0)^2}
+ \frac{A_i(1-A_j) Y_i  Y_j}{(p_i^0)^2 (1 - p_j^0)^2} \\
&\quad + \frac{ (1-A_i)A_j  Y_i Y_j}{(1 - p_i^0)^2 (p_j^0)^2}
+ \frac{(1-A_i)(1-A_j) Y_i  Y_j}{(1 - p_i^0)^2(1 - p_j^0)^2} \Bigr)   \bigg]    \end{align*}     

\subsection{p is a RV - function of covariates V}
\subsubsection{Variance}
$\rho(V_i) = P(A=1 | V_i)$
\begin{align*}
     \mathbb{V}[X_T | V]  &=
    \frac{1}{n^2} \bigg [   \sum_{i=1}^N    \bigl ( \frac{\E[Y_i(1)^2 | V_i]}{\rho(V_i)} + \frac{\E[Y_i(0)^2| V_i]}{1-\rho(V_i)}\bigr)  - n\mathbb{\E} [Y(1) - Y(0) | V_i ]^2
    \\
    &+ 2 \sum_{i,j \in S} \rho_{ij} \cdot \Bigl(
\frac{\mathbb{\E}[Y_i(1) | V_i ] \cdot \mathbb{\E}[Y_j(1)] | V_j]}{\rho(V_i) \rho(V_j)}
+ \frac{\mathbb{\E}[Y_i(1) | V_i ] \cdot \mathbb{\E}[Y_j(0) | V_j]]}{\rho(V_i) (1 -\rho(V_j))} \\
&\quad + \frac{\mathbb{\E}[Y_i(0 | V_i )] \cdot \mathbb{\E}[Y_j(1) | V_j] ]}{(1 - \rho(V_i)) \rho(V_j)}
+ \frac{\mathbb{\E}[Y_i(0 | V_i )] \cdot \mathbb{\E}[Y_j(0) | V_j]}{(1 - \rho(V_i))(1 - \rho(V_j)} \Bigr)   \bigg]
    \end{align*}

\subsubsection{Estimated Variance (IPW)}

\begin{align*}
\mathbb{V}[X_T \mid V]  
&=
\frac{1}{n^2}
\Biggl[
  \sum_{i=1}^N
  \Bigl(
    \frac{A_i\, (Y_i)^2}{\bigl(\rho(V_i)\bigr)^2}
    \;+\;
    \frac{\bigl(1 - A_i\bigr)\, (Y_i)^2}{\bigl(1 - \rho(V_i)\bigr)^2}
  \Bigr)
  \;-\;
  n \,\hat{\tau}(V)^2
\\[6pt]
&\qquad
  +\;
  2 \sum_{i,j \in S}
  \rho_{ij} (V_i, V_j)
  \;\times\;
  \Bigl(
    \frac{A_i\,A_j\,Y_i\,Y_j}{\bigl(\rho(V_i)\bigr)^2\,\bigl(\rho(V_j)\bigr)^2}
    \;+\;
    \frac{A_i\,(1 - A_j)\,Y_i\,Y_j}{\bigl(\rho(V_i)\bigr)^2\,\bigl(1 - \rho(V_j)\bigr)^2}
\\[6pt]
&\qquad\qquad\qquad
    +\;
    \frac{\bigl(1 - A_i\bigr)\,A_j\,Y_i\,Y_j}{\bigl(1 - \rho(V_i)\bigr)^2\,\bigl(\rho(V_j)\bigr)^2}
    \;+\;
    \frac{\bigl(1 - A_i\bigr)\,\bigl(1 - A_j\bigr)\,Y_i\,Y_j}{\bigl(1 - \rho(V_i)\bigr)^2\,\bigl(1 - \rho(V_j)\bigr)^2}
  \Bigr)
\Biggr].
\end{align*}
\newpage

\section{Asymptotic Normality of Estimator}
 Define
\[
x_{t}
\;=\;
\frac{1}{n}\sum_{i=1}^{n}\left(
\frac{p_{i}^{(t)}}{\,p_{i}^{(0)}\,}\,Y_{i}(1)
\;-\;
\frac{1 - p_{i}^{(t)}}{\,1 - p_{i}^{(0)}\,}\,Y_{i}(0)
\right),
\]
where \(x_{0}\) is the estimator using the initial fractional assignments and \(x_{T} = \tau_{\mathrm{swap}}\) is the final binary estimator. Since 
\(\mathbb{E}[\,x_{t}\mid x_{t-1}\,] = x_{t-1}\), 
the sequence \(\{x_{t}\}_{t=0}^{T}\) forms a martingale. Let 
\[
\Delta_{t} \;=\; x_{t} \;-\; x_{t-1}.
\]

\subsection*{Assumptions}

Assume that:
\begin{itemize}
    \item \textbf{Super-Population Assumption:} The units \(\bigl(Y_{i}(1),\,Y_{i}(0),\,V_{i}\bigr)\) are i.i.d.\ draws from a large (hypothetically infinite) population.
    \item \textbf{Bounded Moments:} \(Y_{i}(1)\) and \(Y_{i}(0)\) have uniformly bounded second moments by a constant \(M < \infty\).
    \item \textbf{Bounded Probabilities:} There exists \(\varepsilon > 0\) such that 
    \[
        \varepsilon \;\le\; p_{i}^{(0)} \;\le\; 1 - \varepsilon
        \quad\text{for all }i.
    \]
    \item \textbf{Lower Bound on Quadratic Variation:} There exists a constant \(c > 0\) such that for all sufficiently large \(n\),
    \[
        \bigl\langle x \bigr\rangle_{T}
        \;=\;
        \frac{1}{n}\sum_{t=1}^{T}\mathbb{E}\bigl[\Delta_{t}^{2}\mid x_{t-1}\bigr]
        \;\ge\; c\,.
    \]
\end{itemize}

\subsection*{Quadratic Variation and Lindeberg Condition}

By a law of large numbers for martingale difference arrays under our boundedness assumptions, we have
\[
    \bigl\langle x \bigr\rangle_{T}
    \; \xrightarrow{\;P\;} \; \sigma^{2} \;>\; 0,
\]
with \(\sigma^{2}\) as specified in Equation (6) of the paper. Moreover, since each \(\Delta_{t}\) is bounded (because of bounded potential outcomes combined with bounded probabilities),
\[
    \frac{1}{n}\sum_{t=1}^{T}
    \mathbb{E}\bigl[\Delta_{t}^{2}\,\mathbf{1}_{\{\,|\Delta_{t}|> \varepsilon\sqrt{n}\,\}}
    \mid x_{t-1}\bigr]
    \;\longrightarrow\; 0,
\]
so the Lindeberg condition holds.

\subsection*{Martingale CLT Application}

By the Martingale Central Limit Theorem,
\[
    \sqrt{n}\,\bigl(x_{T} - x_{0}\bigr)
    \;\xrightarrow{d}\; \mathcal{N}\bigl(0,\sigma^{2}\bigr).
\]
Since \(x_{0}\) is the estimator using the initial fractional assignments and \(x_{T} = \tau_{\mathrm{swap}}\), it follows that
\[
    \sqrt{n}\,\bigl(\tau_{\mathrm{swap}} - \tau\bigr)
    \;\xrightarrow{d}\;\mathcal{N}\bigl(0,\sigma^{2}\bigr).
\]
\newpage
\section{Proof of Linear Estimator Unbiasedness and Variance Reduction}
\begin{proof}
Since swap rounding is constructed so that the marginal probability of each treatment indicator is preserved, we have
\[
\E[A_i] = p_i \quad \text{for all } i.
\]
Therefore, by linearity of expectation,
\[
\E[\psi_{swap}] = \E\Biggl[\sum_{i=1}^n a_i A_i + b\Biggr]
= \sum_{i=1}^n a_i\,\E[A_i] + b
= \sum_{i=1}^n a_i\,p_i + b.
\]

Next, to analyze the variance of $\psi_{swap}$, note that since $\psi_{swap}$ is a linear function of the \(A_i\), its variance is given by
\[
\mathbb{V}(\psi_{swap}) = \mathbb{V}\Biggl(\sum_{i=1}^n a_i A_i\Biggr)
= \sum_{i=1}^n a_i^2\,\mathbb{V}(A_i) + 2\sum_{i<j} a_i\,a_j\,\cov(A_i,A_j).
\]
The swap rounding procedure guarantees that for \(i \neq j\),
\[
\cov(A_i, A_j) \le 0.
\]
Since we are now assuming that each \(a_i\) is non-negative, it follows that each product \(a_i\,a_j\) is non-negative. Hence, the cross-covariance terms satisfy
\[
2\sum_{i<j} a_i\,a_j\,\cov(A_i,A_j) \le 0.
\]
It immediately follows that
\[
\mathbb{V}(\psi_{swap}) \le \sum_{i=1}^n a_i^2\,\mathbb{V}(A_i).
\]
This upper bound is exactly the variance one would obtain if the \(A_i\) were independent (since then all covariances vanish). Thus, the variance under swap rounding is no larger than that under independent assignment.
\end{proof}

\newpage
\section*{Proof of Variance Reduction via Covariate-Ordered Pairing under a Bi-Lipschitz Assumption}

For each unit \(i\), define the effective weight
\[
M_i \equiv M(V_i) = \frac{f(V_i)+\tau(V_i)}{p(V_i)} + \frac{f(V_i)}{1-p(V_i)},
\]
and for any pair \((i,j)\) let
\[
\overline{M}_{ij} = \frac{M_i+M_j}{2}.
\]
It is straightforward to verify that
\begin{equation}
M_i M_j = \overline{M}_{ij}^2 - \left(\frac{M_i-M_j}{2}\right)^2.
\label{eq:product_identity}
\end{equation}
Thus, for a fixed pair-average \(\overline{M}_{ij}\), the product \(M_iM_j\) is larger when the absolute difference \(|M_i-M_j|\) is smaller.

To compare two pairing schemes, we impose the following assumptions:

\begin{enumerate}
    \item[\textbf{(A1)}] \textbf{Bi-Lipschitz Condition on \(M(\cdot)\):} Assume that the mapping
    \[
    M: v \mapsto \frac{f(v)+\tau(v)}{p(v)} + \frac{f(v)}{1-p(v)}
    \]
    is bi-Lipschitz on the support of the covariates. That is, there exist constants \(\ell_M > 0\) and \(L_M > 0\) such that for all \(v,v'\),
    \begin{equation}
    \ell_M\,\|v-v'\| \le |M(v)-M(v')| \le L_M\,\|v-v'\|.
    \label{eq:bi_lip}
    \end{equation}
    
    \item[\textbf{(A2)}] \textbf{Assumptions on Pairing Distances:}
    \begin{enumerate}
        \item[(i)] In an \emph{optimal (covariate-ordered) pairing} \(S_{\mathrm{opt}}\), assume that every pair \((i,j)\) satisfies
        \begin{equation}
        \|V_i-V_j\| \le \delta,
        \label{eq:delta_opt}
        \end{equation}
        for some small \(\delta > 0\).
        
        \item[(ii)] In an \emph{arbitrary (random) pairing} \(S_{\mathrm{rand}}\), assume that the expected squared distance between paired covariates is bounded below:
        \begin{equation}
        \mathbb{E}_{(i,j)\sim S_{\mathrm{rand}}}\Bigl[\|V_i-V_j\|^2\Bigr] \ge c^2,
        \label{eq:rand_bound}
        \end{equation}
        for some constant \(c > 0\) (with the natural requirement that \(c > \delta\)).
    \end{enumerate}
\end{enumerate}

Our goal is to show that under these assumptions the differences \(|M_i-M_j|\) in the optimal pairing are uniformly smaller than those in an arbitrary pairing. Consequently, for a fixed pair-average, the product \(M_i M_j\) (and thus the negative covariance contributions in the variance of the estimator) is more favorable under optimal pairing.

\subsection*{Step 1: Upper Bound under Optimal Pairing}

For any pair \((i,j) \in S_{\mathrm{opt}}\), by the upper Lipschitz bound in \eqref{eq:bi_lip} and \eqref{eq:delta_opt} we have
\[
|M_i-M_j| \le L_M\,\|V_i-V_j\| \le L_M\,\delta.
\]
Define
\[
\Delta_{\mathrm{opt}} \equiv L_M\,\delta.
\]
Thus, for every optimally paired \((i,j)\) it holds that
\[
|M_i-M_j| \le \Delta_{\mathrm{opt}}.
\]

\subsection*{Step 2: Lower Bound under Arbitrary Pairing}

For any pair \((i,j)\) drawn according to the arbitrary pairing \(S_{\mathrm{rand}}\), the lower Lipschitz condition in \eqref{eq:bi_lip} implies
\[
(M_i-M_j)^2 \ge \ell_M^2\,\|V_i-V_j\|^2.
\]
Taking expectations over the random pairing, we obtain
\[
\Delta^2_{\mathrm{rand}} \equiv \mathbb{E}_{(i,j)\sim S_{\mathrm{rand}}}\Bigl[(M_i-M_j)^2\Bigr] \ge \ell_M^2\, \mathbb{E}_{(i,j)\sim S_{\mathrm{rand}}}\Bigl[\|V_i-V_j\|^2\Bigr] \ge \ell_M^2\, c^2.
\]
Define
\[
\Delta_{\mathrm{rand}} \equiv \sqrt{\Delta^2_{\mathrm{rand}}} \ge \ell_M\, c.
\]

\subsection*{Step 3: Comparison of \(\Delta_{\mathrm{opt}}\) and \(\Delta_{\mathrm{rand}}\)}

If
\begin{equation}
L_M\,\delta < \ell_M\, c,
\label{eq:comparison}
\end{equation}
then it follows that
\[
\Delta_{\mathrm{opt}} < \Delta_{\mathrm{rand}}.
\]
In other words, the differences \(|M_i-M_j|\) in the optimal pairing are uniformly smaller than the expected differences under an arbitrary pairing.

\subsection*{Step 4: Implication for the Product \(M_iM_j\)}

Recall the identity in \eqref{eq:product_identity}:
\[
M_i M_j = \overline{M}_{ij}^2 - \left(\frac{M_i-M_j}{2}\right)^2.
\]
For a fixed pair-average \(\overline{M}_{ij}\), the product \(M_iM_j\) is maximized when \(|M_i-M_j|\) is minimized. Hence:
\begin{itemize}
    \item Under the optimal pairing, for every pair \((i,j)\),
    \[
    M_i M_j \ge \overline{M}_{ij}^2 - \frac{1}{4}\Delta_{\mathrm{opt}}^2.
    \]
    \item Under arbitrary pairing, the average product satisfies
    \[
    \mathbb{E}_{S_{\mathrm{rand}}}[M_iM_j] \le \mathbb{E}\Bigl[\overline{M}_{ij}^2\Bigr] - \frac{1}{4}\Delta_{\mathrm{rand}}^2.
    \]
\end{itemize}
Since \(\Delta_{\mathrm{rand}}^2 > \Delta_{\mathrm{opt}}^2\) by \eqref{eq:comparison}, the optimal pairing yields, on average, a higher value of \(M_iM_j\).

\subsection*{Step 5: Connection to Variance Reduction}

In our setting, the covariance contribution to the variance of an inverse propensity weighted (IPW) estimator is proportional to
\[
\mathrm{Cov}(X_i,X_j) = \rho_{ij}\,M_i\,M_j,
\]
where \(\rho_{ij} < 0\) is induced by the swap rounding procedure. A larger product \(M_iM_j\) (in absolute value) means that the negative covariance term is more strongly negative. In turn, this leads to a reduction in the overall variance of the estimator.

\subsection*{Conclusion}

Under the modified assumptions---namely, that the effective weight function \(M(\cdot)\) is bi-Lipschitz (with constants \(\ell_M\) and \(L_M\)) and that arbitrary pairings satisfy
\[
\mathbb{E}\Bigl[\|V_i-V_j\|^2\Bigr] \ge c^2,
\]
while the optimal pairing enforces \(\|V_i-V_j\| \le \delta\)---we obtain
\[
\Delta_{\mathrm{opt}} = L_M\,\delta < \ell_M\, c \le \Delta_{\mathrm{rand}}.
\]
Consequently, the product \(M_iM_j\) is higher on average under the optimal (covariate-ordered) pairing than under an arbitrary pairing. This implies that the negative covariance term \(\rho_{ij}\,M_iM_j\) is more pronounced, leading directly to a lower overall variance of the IPW estimator.

\end{document}